\def\year{2019}\relax
\documentclass[letterpaper]{article} 
\usepackage{aaai19}  
\usepackage{amsmath}
\usepackage{amssymb}
\usepackage{dsfont}
\usepackage{boldline}
\usepackage{booktabs}
\usepackage{gensymb}
\usepackage{times}  
\usepackage{helvet}  
\usepackage{courier}  
\usepackage{url}  
\usepackage{graphicx}  
\usepackage{multirow}
\def\etal{\emph{et al.}}
\def\dui{\checkmark}

\begin{document}
%
\title{Joint Unsupervised Learning of Optical Flow and Depth by Watching Stereo Videos}

\author{Yang Wang$^1$ \hspace{0.2cm}  Zhenheng Yang$^2$  \hspace{0.2cm}  Peng Wang$^1$ \hspace{0.2cm} Yi Yang$^1$  \hspace{0.2cm} Chenxu Luo$^3$  \hspace{0.2cm} Wei Xu$^{1}$\\
$^1$Baidu Research \hspace{0.2cm} $^2$ University of Southern California $^3$Johns Hopkins University\\
{\tt\small \{wangyang59, wangpeng54, yangyi05, wei.xu\}@baidu.com  zhenheny@usc.edu   chenxuluo@jhu.edu}}
\maketitle
\begin{abstract}
Learning depth and optical flow via deep neural networks by watching videos has made significant progress recently. In this paper, we jointly solve the two tasks by exploiting the underlying geometric rules within stereo videos. Specifically, given two consecutive stereo image pairs from a video, we first estimate depth, camera ego-motion and optical flow from three neural networks. Then the whole scene is decomposed into moving foreground and static background by comparing the estimated optical flow and rigid flow derived from the depth and ego-motion. We propose a novel consistency loss to let the optical flow learn from the more accurate rigid flow in static regions. We also design a rigid alignment module which helps refine ego-motion estimation by using the estimated depth and optical flow. Experiments on the KITTI dataset show that our results significantly outperform other state-of-the-art algorithms. Source codes can be found at \url{https://github.com/baidu-research/UnDepthflow}
\end{abstract}

\section{Introduction}
Learning 3D scene geometry and scene flow from videos is an important problem in computer vision. It has numerous applications in different areas, including autonomous driving \cite{menze2015object}, robot navigation \cite{desouza2002vision} and video analysis \cite{tsai2016video}. However, collecting ground truths for these tasks could be difficult.

Lots of efforts and progresses have been made recently in unsupervised learning of depth \cite{zhou2017unsupervised} and optical flow \cite{ren2017unsupervised} using neural network based methods. Both approaches have their own advantages and limitations. The depth approach exploits the geometric structure of the scene and decomposes the problem into two orthogonal ones. It can also leverage more frames in time and/or stereo information to add more constraints into the solution \cite{li2017undeepvo}. However, it assumes the entire scene is static and thus has difficulty dealing with moving objects. On the other hand, the optical flow approach can handle moving objects in principle. But it has difficulty in the region of complex structures and occluded areas.  


\begin{figure}
\begin{center}
   \includegraphics[width=1.0\linewidth]{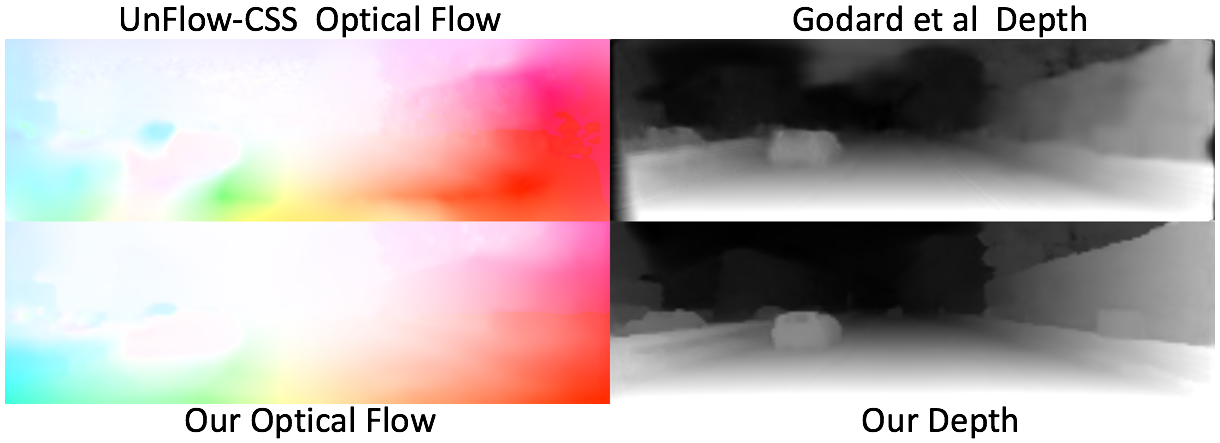}
\end{center}
   \caption{Visualization of the comparison between our method and state-of-the-art unsupervised methods \cite{Meister2018UUL,godard2016unsupervised}. It can be seen that our method produces smoother results in planes and sharper boundaries. }
\label{fig:comp}
\end{figure}

Based on the above observation, we propose a framework to jointly learn depth and optical flow by mutually leveraging the advantages from each other. We first estimate the moving region by comparing the optical flow and the rigid flow induced by the camera motion. With the estimated moving region mask, depth and camera motion estimations can be improved by not considering the reconstruction loss in the moving region. On the other hand, the optical flow performance can also be improved by learning from the more accurate rigid flow in the static region (see Fig.\ref{fig:comp} for an example). 

Some works \cite{ranjan2018adversarial,yang2018every,yin2018geonet} have tried to learn the two tasks jointly with monocular videos. We however decide to use stereo pairs as input to our framework for the following reasons. On one hand, the performance of the unsupervised depth estimation is much better with stereo input compared to monocular-based methods \cite{godard2016unsupervised}. Given the fact that stereo cameras have a wide range applications in real world scenarios (e.g. on smartphones), we feel that unsupervised learning of depth with stereo pairs itself is a topic worth studying. Additionally, with stereo inputs both depth and visual odometry can recover the absolute scale which could be more useful in certain tasks like self-localization.

On the other hand, the higher quality stereo-based depth can be utilized to improve estimations of other quantities with our newly designed modules. It has been noticed that directly estimating camera motion from two images can be difficult, because it would implicitly require the network to also estimate the scene structure \cite{wang2018learning}. Based on the high quality stereo-based depth, we propose a rigid alignment module to facilitate the camera motion estimation. More concretely, we refine the camera motion by rigidly aligning two consecutive point clouds whose correspondences are established through optical flow. Additionally, with the high quality depth and better ego-motion estimations, the rigid flow derived from ego-motion is much more accurate than monocular-based ones. Therefore we propose a flow consistency module which lets the optical flow learn from the rigid flow in the static regions. 

In summary, the key contributions of this work are: 1) a unified framework for unsupervised learning of optical flow, depth, visual odometry and motion segmentation with stereo videos. 2) a rigid alignment module for refining the ego-motion estimation. 3) a flow consistency module for learning optical flow from rigid flow. Our method improves the state-of-the-art performance of unsupervised learning of depth and optical flow by a large margin. On KITTI 2012, for example, our method reduces the optical flow error from previous state-of-the-art unsupervised method \cite{Meister2018UUL} by 50\%, and reaches the performance of the supervised methods.

\section{Related Work}

\subsection{Unsupervised Learning of Depth and Ego-motion}
The unsupervised learning of depth and ego-motion through monocular videos using deep learning was first achieved in \cite{zhou2017unsupervised}. Later, different methods were proposed to improve the results. \cite{yang2017unsupervised} added a depth-normal consistency term. To make the visual odometry more accurate, \cite{wang2018learning} used direct method to refine the pose estimation, and \cite{mahjourian2018unsupervised} proposed a 3D ICP loss. Our rigid alignment module tries to tackle the same issue and is similar to \cite{mahjourian2018unsupervised}. The difference is that we use optical flow to find the point cloud correspondence in a single pass while they used nearest neighbor method to iteratively refine the correspondences which would require longer time. 

Another line of work used stereo images to learn the depth where the relative pose of the camera is fixed and known \cite{garg2016unsupervised}. \cite{godard2016unsupervised} introduced a left-right consistency loss for improvement. \cite{zhan2018unsupervised} and \cite{li2017undeepvo} combined stereo pairs and monocular video matching together. The aforementioned works all use a monocular image as input to estimate the depth although stereo images are needed at the training time. \cite{zhou2017unsupervisedstereo,godard2016unsupervised} used stereo pairs as input to estimate the depth which is the same as our approach, and have much better performance compared to monocular input methods.

\subsection{Unsupervised Learning of Optical Flow}
The unsupervised learning of optical flow with a neural network was first introduced in \cite{ren2017unsupervised} and \cite{jason2016back}. Later, \cite{wang2018occlusion} and \cite{Meister2018UUL} improved the results by explicitly handling the occlusions. However, there is still a gap for the unsupervised learning methods to reach the performance of the supervised methods \cite{ilg2016flownet,sun2017pwc}. 

\subsection{Joint Learning of Depth and Flow}
There are a large body of works on scene flow estimation using traditional variational-based methods \cite{vedula1999three}. Here we only list some of the most recent ones. \cite{menze2015object} modeled the whole scene as a collection of piece-wise planes based on superpixels with rigid motions. \cite{behl2017bounding} improved upon it by adding recognition and instance segmentation information. \cite{wulff2017optical} segmented the scene into static and moving regions where the optical flow in the static region is refined using the plane+parallax method. The framework in \cite{taniai2017fast} is similar to ours, in which depth, ego-motion, optical flow and motion segmentation are optimized together. The major limitation of traditional methods is computation time.

There are also works on combining neural network based unsupervised learning of depth and optical flow. \cite{yin2018geonet} used a residual FlowNet to refine the rigid flow from depth and ego-motion to the full optical flow, but it did not account for the moving objects in the rigid flow estimation. \cite{yang2018every} handled moving objects explicitly with motion segmentation but did not use depth to improve optical flow. \cite{ranjan2018adversarial} pieced the optical flow and rigid flow together to explain the whole scene in a adversarial collaboration. Their method requires iterative training and forwarding all three networks to obtain the optical flow, while our method can be simply trained together and optical flow estimation only depends on one network. 

\section{Method}
We will first give an overview of our method and then describe specific components in details. 

\begin{figure}
\begin{center}
   \includegraphics[width=1.01\linewidth]{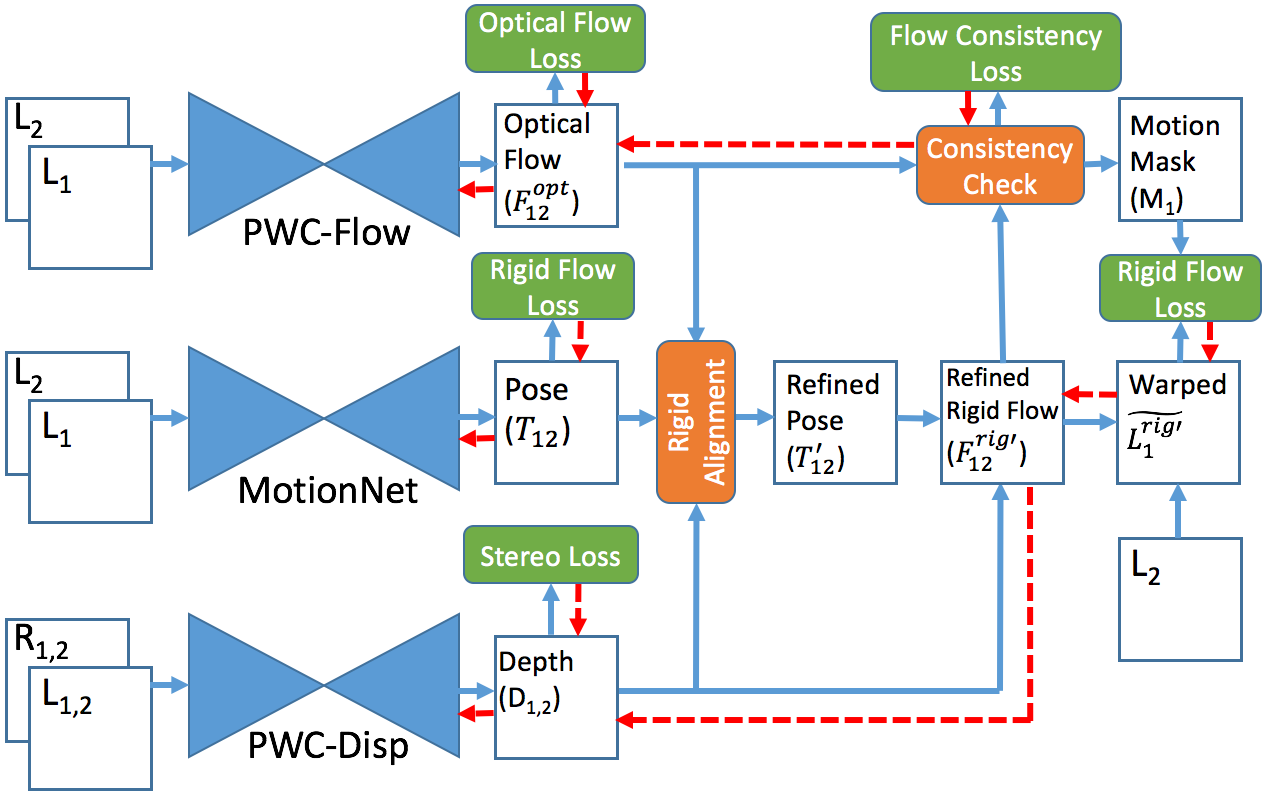}
\end{center}
   \caption{Schematic overview of our method. It learns to estimate depth, optical flow, camera motion, and motion segmentation from two consecutive stereo pairs in an unsupervised manner. The different components of our unsupervised loss are depicted in green boxes. Our proposed rigid alignment module refines the estimated camera pose, and the flow consistency check between optical flow and rigid flow gives the motion segmentation mask. The red arrows indicate the gradient back-propagation directions.}
\label{fig:network}
\end{figure}

\subsection{Overall Structure}
The overall schematic structure of our method is shown in Fig.\ref{fig:network}. During training, the inputs to our system are two consecutive stereo pairs ($L_1$, $R_1$, $L_2$, $R_2$ where $L_1$ denotes the left image at time $t_1$ and so forth) .  It has three major neural network modules: one for estimating optical flow ($F_{12}^{opt}$) between two consecutive left images (PWC-Flow), one for estimating relative camera pose ($T_{12}$) between two consecutive left images (MotionNet), and one for estimating disparity between stereo pair of images (PWC-Disp). With a known stereo baseline $B$ and horizontal focal length $f_x$, The disparity ($d$) can be converted into absolute scale depth $D = B f_x / d$.    

By combining $D_1$ and $T_{12}$, one can calculate the flow induced by the camera motion. We label this flow as rigid flow ($F^{rig}_{12}$) since it assumes that the whole scene is static (this part of calculation graph is not depicted in Fig.\ref{fig:network} due to the space constraint). Then the rigid alignment module will refine the MotionNet estimated pose $T_{12}$ to be $T^{'}_{12}$ by using the optical flow and depths. Similarly, we can get the refined rigid flow ($F^{rig'}_{12}$) by combining $D_1$ and $T^{'}_{12}$. In the next step, we perform a consistency check between $F_{12}^{opt}$ and $F^{rig'}_{12}$. The region is labeled as moving foreground ($M_1$) if the difference between the two flows is greater than a threshold, and the rest of image is labeled as static background. 

The whole system is trained in an unsupervised manner. The unsupervised loss for flow $F_{12}^{opt}$ follows the method in \cite{wang2018occlusion} which consists of an occlusion-aware reconstruction loss and a smoothness loss. The unsupervised loss for depth $D$ follows the method in \cite{godard2016unsupervised} which consists of a reconstruction loss, a smoothness loss and a left-right consistency loss. We also add a reconstruction loss between $L_1$ and $\widetilde{L}_1^{rig}$ ($\widetilde{L}_1^{rig'}$) in the static region, where $\widetilde{L}_1^{rig}$ ($\widetilde{L}_1^{rig'}$) is obtained by warping $L_2$ using $F^{rig}_{12}$ ($F^{rig'}_{12}$). In addition, we have a consistency loss between $F_{12}^{opt}$ and $F^{rig'}_{12}$ in the static region ($1 - M_1$). The directions of gradient back-propagations are labeled as red arrows in Fig. \ref{fig:network}. 

\subsection{Network Architecture Design}
The structure of PWC-Flow follows the design in \cite{sun2017pwc} because of its lightweight and excellent supervised optical flow performance. 

The structure of MotionNet is similar to the one used in \cite{zhou2017unsupervised} except that the input to the network has only two consecutive images instead of three or five. We add more convolutional layers into our MotionNet because we find that it gives better pose estimation when using only two consecutive images. 

Since optical flow and disparity estimations are both problems of finding correspondence and indeed very similar in nature, we modify PWC-net to exploit the special structure of the disparity estimation and turn it into PWC-Disp. PWC-Disp only searches the horizontal direction when calculating the cost volume, only estimates the horizontal component of the flow (forcing the vertical component to be zero), and forces the flow to be negative (so that $\text{disp} = - \text{flow}_{x}/\text{image\_width}$ is always positive).  We choose PWC-Disp as our network to estimate disparity due to its lightweight. 

\begin{figure}
\begin{center}
   \includegraphics[width=0.95\linewidth]{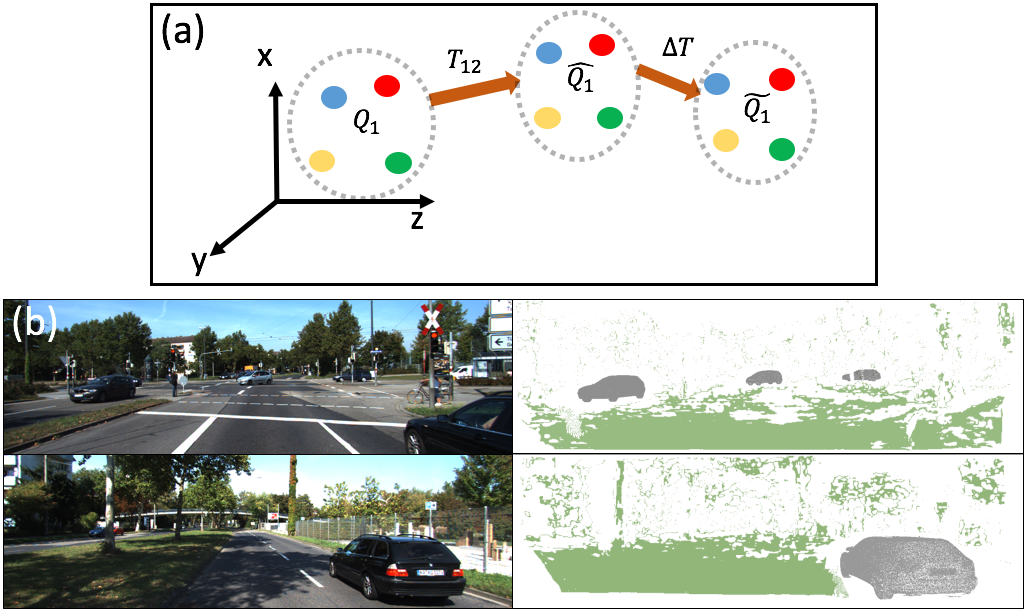}
\end{center}
   \caption{(a) Schematic illustration of the rigid alignment module. Each of the point clouds $Q_1$, $\widehat{Q}_1$ and $\widetilde{Q}_1$ is depicted by colored balls enclosed in a dashed circle. The color of the balls encodes correspondences between point clouds. (b) Visualization of the region chosen for rigid alignment. Left column plots left images ($L_1$). Right column plots the region used in the rigid alignment module (i.e., region $R$ described in Eq. \ref{eqn:deltat}) in green overlaying on the ground truth moving object mask in black.}
\label{fig:align}
\end{figure}


\subsection{Rigid Alignment Module}
During rigid alignment, we first transform the points in 2D image space to 3D point cloud using Eq.\ref{eqn:proj}, where $P_t(i,j)$ is the homogenous coordinate of the pixel at the location of $(i,j)$ of image $L_t$, $K$ is the camera intrinsic matrix, $D_t (i,j)$ is the estimated depth of $L_t$ at location $(i,j)$ and $Q_t (i,j)$ is the corresponding 3D coordinate (i.e., $x, y, z$) of the pixel.  

\begin{equation}
Q_t (i, j) = D_t (i,j) K^{-1} P_t (i, j)
\label{eqn:proj}
\end{equation}

We then transform $Q_1$ into $\widehat{Q}_1 = T_{12} Q_1$, where $T_{12}$ is the initial estimation of the pose from MotionNet, and $\widehat{Q}_1$ is the 3D coordinate of the point in $L_1$ at the time of $t_2$. We can also obtain a similar quantity $\widetilde{Q}_1$ by warping $Q_2$ back to the frame of $t_1$ using optical flow $F^{opt}_{12}$. This is the same bilinear sampling method used to warp the image $L_2$ back to $\widetilde{L_1}$ (see Eq.\ref{eqn:warp}). This warping step is to establish correspondence so that $\widehat{Q}_1 (i, j)$ corresponds to $\widetilde{Q}_1 (i,j)$. 

\footnotesize
\begin{equation}
\begin{split}
\widetilde{Q}_1(i, j) = \sum_{m=1}^{W}\sum_{n=1}^{H} Q_2(m,n) \max (0, 1-|m - (i+F_{12}^{x}(i, j))| ) \\
  \cdot \max (0, 1-|n - (j+F_{12}^{y}(i, j))| )
\end{split}
\label{eqn:warp}
\end{equation}
\normalsize

If everything is perfectly accurate, $\widehat{Q}_1$ should be equal to $\widetilde{Q}_1$ in the static and non-occluded region of the scene. Therefore we can refine our pose estimation by rigidly aligning these two point clouds (a schematic illustration of the rigid alignment module can be seen in Fig. \ref{fig:align}a). Concretely, we estimate the refinement pose $\Delta T$ by minimizing the distance between $\Delta T\widehat{Q}_1$ and $\widetilde{Q}_1$ in selected region $R$:

\begin{equation}
\Delta T  = \operatorname*{argmin}_{\Delta T} \sum_{R} ||\Delta T\widehat{Q}_1 - \widetilde{Q}_1||^2 
\label{eqn:deltat}
\end{equation}

This minimization is performed exactly using the method described in \cite{besl1992method} which involves solving an SVD problem. We choose region $R$ to be 25\% of the points in the non-occluded area that have the smallest distance between $\widehat{Q}_1$ and $\widetilde{Q}_1$. By doing this, we try to exclude points in the moving region since they tend to have larger distances between $\widehat{Q}_1$ and $\widetilde{Q}_1$. Examples of the region R can be seen in Fig. \ref{fig:align}b which shows clear separation between region $R$ (green) and moving objects (black). On the 200 KITTI 2015 training images, only 1.4\% of the moving object pixels fall into the region $R$.  The non-occluded area ($O_1$) is estimated using reverse optical flow ($F_{21}^{opt}$) as described in \cite{wang2018occlusion}. The refined pose can be obtained by combining $T_{12}$ and $\Delta T$:

\begin{equation}
T^{'}_{12} = \Delta T \times T_{12}
\label{eqn:comb}
\end{equation}

\begin{figure}
\begin{center}
   \includegraphics[width=0.95\linewidth]{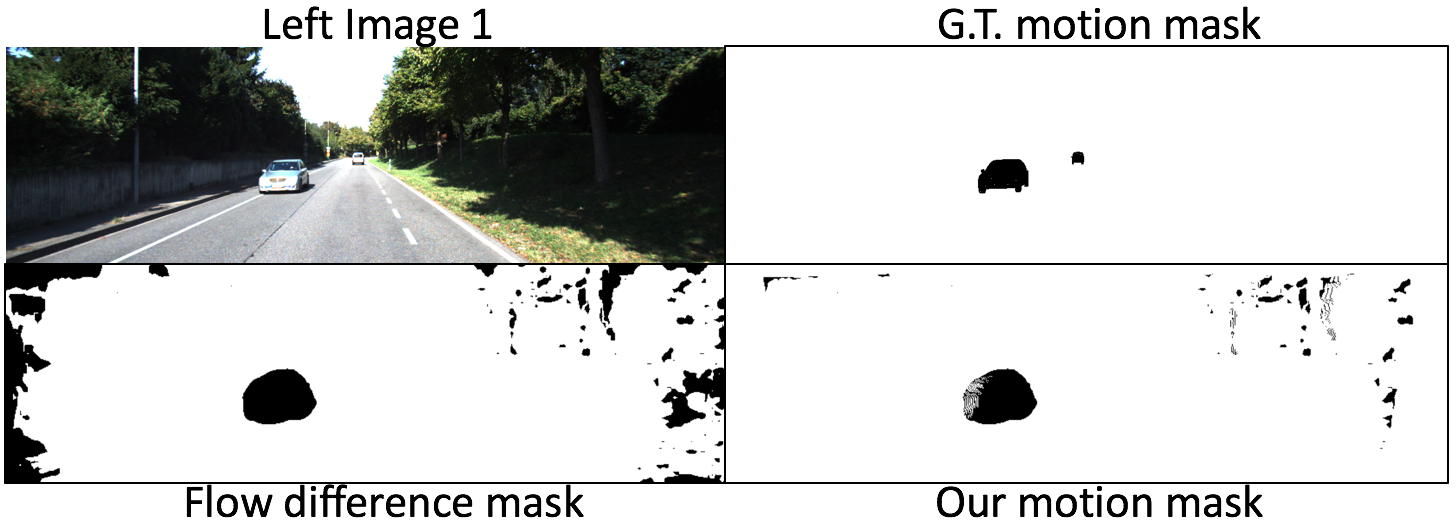}
\end{center}
   \caption{An example of motion mask estimation. After taking away the occluded area, the estimated motion mask (bottom right) improved compared to the mask obtained solely by thresholding the flow difference (bottom left).}
\label{fig:cons}
\end{figure}

\subsection{Flow Consistency and Motion Segmentation}
With the refined pose $T^{'}_{12}$ in hand, we can calculate the rigid flow induced by camera motion to be:

\begin{equation}
F^{rig'}_{12} = K T^{'}_{12} D_1 K^{-1} P_1 - P_1
\label{eqn:comb}
\end{equation}

If $F_{12}^{opt}$ and $F^{rig'}_{12}$ were both accurate, their values should match in the static region and differ in the moving region. Based on this observation, we estimate the moving region to be:

\begin{equation}
M_1 = \mathds{1} (||F^{opt}_{12} - F^{rig'}_{12}|| > \delta \cap O_1) 
\label{eqn:comb}
\end{equation}

We also force the estimated moving region to be in the non-occluded area ($O_1$) because $F^{opt}_{12}$ is less accurate in the occluded area which could lead to false positives (see Fig.\ref{fig:cons} for an example). 

\subsection{Unsupervised Losses}
The unsupervised loss of our method is composed of four components: optical flow loss ($l_{opt}$), stereo loss ($l_{st}$), rigid flow loss ($l_{rig}$), and flow consistency loss ($l_{con}$). We will describe each loss in details below. 

\textbf{Optical Flow Loss} The optical flow loss is similar to the one described in \cite{wang2018occlusion} which has an occlusion-aware reconstruction loss term ($l_{opt-ph}$) and a smoothness loss term ($l_{opt-sm}$).  $l_{opt-ph}$ is a weighted average between the SSIM-based loss and the absolute photometric difference loss on the non-occluded area, where $\widetilde{L}_{1}^{opt}$ is the reconstruction of $L_1$ by warping $L_2$ using $F^{opt}_{12}$. $l_{opt-sm}$ is the average absolute value of the edge-weighted second-order derivative of the optical flow on the moving foreground region. The constraint for the optical flow on the static region will be provided in the consistency loss part.

\scriptsize
\begin{equation*}
\begin{split}
&l_{opt-ph} = \Psi(L_1, \widetilde{L}_{1}^{opt}, O_1) \\
                  &= \frac{1}{\sum_{i, j} O_1} [\sum_{i,j} (\alpha \frac{1 - SSIM(L_1, \widetilde{L}_{1}^{opt})}{2} + (1-\alpha) |L_1 - \widetilde{L}_{1}^{opt}|)  \cdot O_1] \\
&l_{opt-sm} = \frac {1}{N} \sum_{i, j} \sum_{d \in x, y} |\partial_d^2 F^{opt}_{12}(i, j)| e^{-\beta |\partial_d L_1(i, j)|} \cdot M_1
\end{split}
\end{equation*}
\normalsize

\textbf{Stereo Loss} The stereo loss is the same as in \cite{godard2016unsupervised}.

\textbf{Rigid Flow Loss} The rigid flow loss is a reconstruction loss term applied on $\widetilde{L}_{1}^{rig}$ and $\widetilde{L}_{1}^{rig'}$ in the static region. $\widetilde{L}_{1}^{rig}$ ($\widetilde{L}_{1}^{rig'}$) is the reconstruction of $L_1$ by warping $L_2$ using $F_{12}^{rig}$ ($F_{12}^{rig'}$).  

\footnotesize
\begin{equation*}
\begin{split}
&l_{rig}  = l_{rig}^{1} + l_{rig}^{2} \\
& = \Psi(L_1, \widetilde{L}_{1}^{rig}, O_1 \cdot (1-M_1)) + \Psi(L_1, \widetilde{L}_{1}^{rig'}, O_1 \cdot (1-M_1))
\end{split}
\end{equation*}
\normalsize

Here we also include $l_{rig}^{1}$ into the loss because the rigid alignment module is non-differentiable and we need $l_{rig}^{1}$ to supervise the MotionNet.

\textbf{Flow Consistency Loss} From the experimental result below, we find that $F_{12}^{rig'}$ is more accurate than $F_{12}^{opt}$ in the static region. Therefore we decide to use $F_{12}^{rig'}$ to guide the learning of $F_{12}^{opt}$ using the following one-sided consistency loss term, where $SG$ stands for stop-gradient. 

\begin{equation*}
l_{con} = \frac {1}{N} \sum_{i,j} |F_{12}^{opt}(i,j) - SG(F_{12}^{rig'}(i,j))| \cdot (1-M_1(i,j))
\end{equation*}

\textbf{Total Loss} The total loss is a weighted sum of the aforementioned losses:

\begin{equation*}
\begin{split}
l_{total} = &l_{opt-ph} + \lambda_{sm} l_{opt-sm} + \lambda_{st} l_{st} \\
                 &+ \lambda_{rig} l_{rig} + \lambda_{con} l_{con}
\end{split}
\end{equation*}

\begin{figure*}[t]
\begin{center}
   \includegraphics[width=0.9\linewidth]{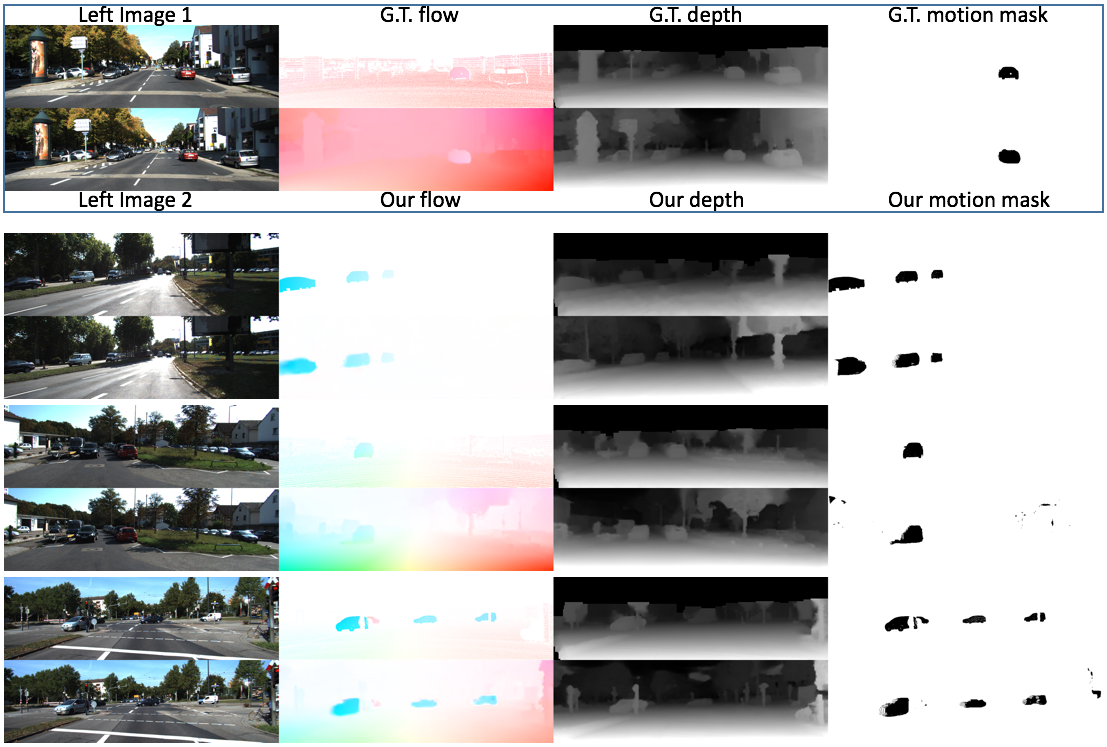}
\end{center}
   \caption{Qualitative results of our method. Each instance of examples is plotted in two rows, including left image 1 ($L_1$), left image 2 ($L_2$), ground truth optical flow, our estimated optical flow ($F^{opt}_{12}$), ground truth depth, our estimated depth ($D_1$), ground truth motion mask, and our estimated motion mask ($M_1$). }
\label{fig:vis}
\end{figure*}

\section{Experiments}

We evaluate our methods on the KITTI dataset, and compare our results to existing supervised and unsupervised methods on the tasks of depth, optical flow, camera motion, scene flow and motion segmentation.

\textbf{Training Details} The whole training process contains three stages. In the first stage, we train the PWC-Flow network using $l_{opt-ph} + \lambda_{sm} l_{opt-sm}$. In the second stage, we then train the PWC-Disp and MotionNet networks using $\lambda_{st} l_{st} + \lambda_{rig} l_{rig}^{1}$ without the rigid alignment and flow consistency check modules (i.e., $M_1$ is set to zero everywhere). This is because the rigid alignment and flow consistency check only work when both depths and optical flows are estimated reasonably accurate. In the third stage, everything is trained together using the total loss $l_{total}$. 

In all of three stages, we used Adam optimizer \cite{kingma2014adam} with $\beta_1 = 0.9$ and $\beta_2 = 0.999$. The learning rate is set to be $10^{-4}$.  The hyper-parameters $[\lambda_{sm}, \lambda_{st}, \lambda_{rig}, \lambda_{con}, \alpha, \beta, \delta]$ are set to be $[10.0, 1.0, 10.0, 0.01, 0.85, 10.0, 3.0]$. 

During training, we use batch size of 4. In each stage, we train for around 15 epochs and choose the model with best validation accuracy for the start of training next stage. Images are scaled to have values between 0 and 1, and size of 832 $\times$ 256. The only data augmentation we perform is random left-right flipping and random time order switching (swapping $t_1$ and $t_2$). 

\textbf{Dataset} For the depth, optical flow, scene flow and motion segmentation tasks, we train our network using all of the raw data in KITTI excluding the scenes appeared in KITTI 2015 \cite{menze2015object} so that we could use the training set of KITTI 2015 as our validation set, and compare with existing methods. None of KITTI 2012 \cite{geiger2012we} data exists in the KITTI raw data so we could also evaluate our model on it. Notice that KITTI 2012 dataset only contains static scenes. For the odometry task, we use sequences 00-08 as training data and sequences 09, 10 as validation data. All of our models are trained in a pure unsupervised manner.

\begin{table*}[t]
\centering
\fontsize{8.0}{10}\selectfont
\begin{tabular}{ l  ccc | c  c  c  c |  c c c c  }
   \hlineB{3}
                 &&&                                        &\multicolumn{4}{c|}{KITTI 2012}       & \multicolumn{4}{c}{KITTI 2015} \\
    Method &Train   & Test    &Super-   & train     & train   &  train    & test      &  train & train   & train    & test \\
                 &Stereo& Stereo &vised     & Noc     & Occ    &  All       & All        & move & static & all       &all  \\
   \hline
   
   Flownet2    &      &          &\checkmark &  --  &    --     & \bf{4.09}     &      --    &    --      &   --     & \bf{10.06}  & -- \\
   Flownet2+ft &      &         &\checkmark&  --    &    --     & (1.28)  &  1.8    &     --     &     --   & (2.3)      & 11.48\% \\
   
   PWC-Net    &      &         &\checkmark&  --    &     --    & 4.14     &     --     &     --     & --       & 10.35  & -- \\
   PWC-Net+ft &      &         &\checkmark&  --    &   --     & (1.45)  &  \bf{1.7}    &      --    &     --   & (2.16)  &  \bf{9.60\%} \\
   
   \hline
   UnFlow-CSS &      &         &                &1.26 &  --       & 3.29    &      --      &       --   &    --  & 8.10    &   --     \\
   Geonet          &      &         &                 &  --   &    --     &    --     &       --     &      --    &   --   & 10.81   &  --      \\
   \cite{ranjan2018adversarial} &      &         &         &   --    & --     &    --    &     --      & 6.35   & 6.16  & 7.76   &    --    \\
   \cite{wang2018occlusion}        &      &        &        &   --      &   --   & 3.55   & 4.2         &      --   &  --    & 8.88     & 31.2\% \\
   Ours (PWC-only) &      &         &           &1.15   & 11.2  &2.68 &     --   & 5.92    &7.68  & 7.88    &    --    \\
   Ours (Ego-motion)&\dui&\dui  &          &2.27   & 6.67   &2.86 &     --   & 35.9  &4.53  & 11.9      &    --   \\
   Ours (Ego+align) &\dui&\dui  &          &1.46     & \bf{4.88}  &1.93 &    --         &  36.5 &\bf{2.99}  & 10.69    & --  \\
   Ours (Full)           &\dui &        &          & \bf{1.04}& 5.18  & \bf{1.64} & \bf{1.8}         & \bf{5.30}   & 5.39 &  \bf{5.58}   & \bf{18.00 \%} \\
   \hline
   Ours (mono-Ego-motion) &\dui&         &          & 2.78  & 8.47 & 3.58  &    --       & 34.8 & 6.56   & 13.5   &   --     \\
  \hlineB{3}  
\end{tabular}
\caption{Quantitative evaluation on the optical flow task. The numbers reported here are all average end-point-error (EPE) except for the last column (KITTI2015 test) which is the percentage of erroneous pixels (Fl-all). A pixel is considered to be correctly estimated if the flow end-point error is \textless3px or \textless5\%. The upper part of the table contains supervised methods and lower part of the table contains unsupervised methods. For all metrics, smaller is better. The best results from each category are boldfaced. }
\label{tab:op}
\end{table*}

\begin{table*}
\vspace{-0.5\baselineskip}
\centering
\fontsize{8.0}{9}\selectfont
\def\arraystretch{1.15}
\begin{tabular}{l|ccc|ccccc|ccc}
\specialrule{.2em}{.1em}{.1em}
\multirow{2}{*}{Method}                              & Train    & Test       &Super-&      \multicolumn{5}{c|}{Lower the better}             &            \multicolumn{3}{c}{Higher the better}      \\                                                               
                                        	                          & Stereo &  Stereo  &vised&  Abs Rel    & Sq Rel    & RMSE  & RMSE log   & D1-all & $\delta < 1.25$ & $\delta < 1.25^2$ & $\delta < 1.25^3$ \\ 
\hline
\cite{zhou2017unsupervised}       &            &              &         &      0.216     & 2.255      & 7.422   &     0.299      &  -- &         0.686         & 0.873                   & 0.951                    \\
DDVO                                       &              &              &         &    0.151       & 1.257      & \bf{5.583}   & 0.228          & --  &         0.810         & 0.936                   & 0.974                     \\
(Godard et al. 2017)                   &\checkmark  &     &         &    0.124      & 1.388     & 6.125   &  0.217  & 30.27\%&         0.841         & 0.936                  & 0.975                      \\
\cite{yang2018every}                                   & \checkmark  &     &         &     \bf{0.109} & \bf{1.004} & 6.232    & \bf{0.203}            &  -- &    \bf{0.853}     & \bf{0.937}         & \bf{0.975}                      \\ 

\hline
(Godard et al. 2017)                  &\checkmark& \checkmark&&  0.068      & 0.835     & 4.392   &  0.146  & 9.194\%  &   0.942    & 0.978        & 0.989                      \\
Ours (Stereo-only)                      &\checkmark& \checkmark&&  0.060      & 0.833     & 4.187   &  0.135  & 7.073\%  &   0.955    & 0.981        & 0.990                      \\
Ours (Ego-motion)                &\checkmark& \checkmark&&  0.052      & 0.593     & 3.488   &  0.121  & 6.431\%  &   0.964    & \bf{0.985}        & 0.992                      \\
Ours (Full)           			&\checkmark& \checkmark&&\bf{0.049}  & \bf{0.515} & \bf{3.404} & \bf{0.121} & 5.943\%  & \bf{0.965}  & 0.984   &  \bf{0.992}     \\
\hline
PSMNet                                     &\checkmark& \checkmark& \checkmark&  --   &   --     &     --   &   --   & \bf{1.83\%}    &    --      &         --     &       --               \\
\hline

\end{tabular}
\caption{Quantitative evaluation of the depth task on the KITTI2015 training set. Since depths and disparities are directly related to each other, we put them into the same table. Abs Rel, Sq Rel, RMSE, RMSE log, $\delta < 1.25, 1.25^2, 1.25^3$ are standard metrics for depth evaluation. We capped the depth to be between 0-80 meters to compare with existing literatures. D1-all is the error rate of the disparity. }
\label{tab:dep}
\end{table*}

\textbf{Optical Flow Evaluation} We evaluate our method on the optical flow estimation task using both KITTI 2012 and KITTI 2015, and the quantitative results are shown in Table \ref{tab:op}. Ours (PWC-only) is our baseline model after training PWC-Flow using only the loss $l_{opt}$ in the first stage. We could see that it is already better than state-of-the-art unsupervised optical flow method UnFlow-CSS \cite{Meister2018UUL} demonstrating the effectiveness of our occlusion-aware loss and PWC network structure. 

Ours (Ego-motion) is the result of rigid flow $F^{rig}_{12}$ at the end of the second stage training. The rigid flow is shown to be better than the previous general optical flow in occluded (6.67 vs. 11.2) and static (4.53 vs. 7.68) regions. This observation is consistent with our assumption about the advantage of the rigid flow in those areas, and provides motivation for our proposed flow consistency loss ($l_{con}$). The rigid flow is much worse in moving regions which is expected since it is only supposed to be accurate in static regions. 

Ours (Ego+align) is the result of refined rigid flow $F^{rig'}_{12}$ at the beginning of the third stage training (i.e., its only difference from Ours (Ego-motion) is applying the rigid alignment module without any changes to the network parameters). The result shows that the rigid alignment module significantly improves the rigid flow in static regions (1.93 vs. 2.86 and 2.99 vs. 4.53). 

By guiding the optical flow $F_{12}^{opt}$ using $F^{rig'}_{12}$ in static regions, we reach our final model Ours (Full) for optical flow estimation at the end of the third training stage. It is still worse than $F^{rig'}_{12}$ in static regions but has the best overall performance. For KITTI 2012, our method reduces the error from previous state-of-the-art unsupervised method \cite{Meister2018UUL} by 50\%, and reaches the performance of the supervised methods, which demonstrates the benefits of our proposed method and the utilization of stereo data. For KITTI 2015, our method also outperforms previous unsupervised methods by a large margin, but still falls behind supervised methods. Although our method needs extra stereo data during training compared to previous methods, it only requires two consecutive monocular images during the test time. Qualitative examples of our estimated optical flow can be seen in Fig. \ref{fig:vis}.

We also conduct an ablation study using only monocular image as input to the depth network and follow the same training process as Ours (Ego-motion). The evaluation result is shown as Ours (mono-Ego-motion). We can see that the monocular model produce much worse rigid flow compared to the stereo model, and thus would not provide as much benefits for guiding the optical flow learning.  

\begin{table}
\centering
\fontsize{7.2}{10}\selectfont
\begin{tabular}{ c | c c | c  c }
   \hlineB{3}
    Method                  & frames & Stereo &    Sequence 09         & Sequence 10  \\
    \hline
    ORB-SLAM(Full)   &  All                 &              &  0.014 $\pm$ 0.008  &  0.012 $\pm$ 0.011 \\ 
    \cite{zhou2017unsupervised} &  5  &              &  0.016 $\pm$ 0.009  &  0.013 $\pm$ 0.009 \\          
    Geonet             &  5                 &              &  0.012 $\pm$ 0.007  &  0.012 $\pm$ 0.009 \\   
    Mahjourian \etal  &  3  & &  0.013 $\pm$ 0.010  &  0.012 $\pm$ 0.011 \\ 
    \cite{ranjan2018adversarial}            & 5 &   &  0.012 $\pm$ 0.007   & \bf{0.012 $\pm$ 0.008}  \\
    
    \hline
    Ours (Ego-motion) &  2           &   \dui   &  0.023 $\pm$ 0.010  &  0.022 $\pm$ 0.016 \\
    Ours (Ego+align) &  2           &   \dui   &  0.013 $\pm$ 0.006  &  0.015 $\pm$ 0.010 \\ 
    Ours (Full) &  2              &   \dui   &  \bf{0.012 $\pm$ 0.006}  & 0.013 $\pm$ 0.008  \\ 
  \hlineB{3}  
\end{tabular}
\caption{Quantitative evaluation of the odometry task using the metric of the absolute trajectory error. }
\label{tab:ate}

\quad
\centering
\fontsize{7.2}{10}\selectfont
\begin{tabular}{ c | c c | c c c c }
   \hlineB{3}
    \multirow{2}{*}{Method}    &    \multicolumn{2}{c|}{Sequence 09}  & \multicolumn{2}{c}{Sequence 10}  \\
                                            &  $t_{err}\%$   &  $r_{err}(\degree / 100)$ &  $t_{err}\%$   &  $r_{err}(\degree / 100)$ \\
    \hline
    ORB-SLAM(Full)              &  15.30  & \bf{0.26} & \bf{3.68} & \bf{0.48} \\ 
    \cite{zhan2018unsupervised}  & 11.92     & 3.60  & 12.62 & 3.43 \\
    
    Ours (Ego-motion)          &  13.98  & 5.36 & 19.67 & 9.13   \\
    Ours (Ego+align)            &  8.15  &  3.02 & 9.54 & 4.80  \\ 
    Ours (Full)                       &  \bf{5.21} & 1.80   & 5.20 & 2.18    \\ 
  \hlineB{3}  
\end{tabular}
\caption{Quantitative evaluation of the odometry task using the metric of average translational and rotational errors. Numbers of ORB-SLAM (Full) are adopted from \cite{zhan2018unsupervised}.}
\label{tab:odo}
\end{table}

\textbf{Depth Evaluation} We evaluate our depth estimation on the KITTI 2015 training set and the results are shown in Table \ref{tab:dep}. Ours (Stereo-only) is the PWC-Disp network trained using only stereo images and the loss $l_{st}$. This is not part of our three training stages but only serves as an ablation study. Ours (Stereo-only) is already better than the best stereo-based unsupervised learning method \cite{godard2016unsupervised} demonstrating the effectiveness of our PWC-Disp network. We can observe that stereo-based methods still have much better performance than monocular methods. 

Ours (Ego-motion) is the model at the end of our second training stage. After adding the data of time consecutive images into the training, the depth accuracy improves, especially in the large distance regions (0.593 vs. 0.833). 

Ours (Full) model improves further after adding the rigid alignment module and explicitly handling the moving regions. However, its performance is still relatively far away from the supervised method like PSMNet \cite{chang2018pyramid}. The depth and scene flow evaluation on KITTI 2015 test set can be found in the supplementary materials. The qualitative results of our estimated depth can be seen in Fig. \ref{fig:vis}. 

\textbf{Odometry Evaluation} We evaluate our odometry performance using two existing metrics. One metric is proposed in \cite{zhou2017unsupervised} which measures the absolute trajectory error averaged over all overlapping 5-frame snippets after a single factor rescaling with the ground truth. The results for this metric are shown in Table. \ref{tab:ate}. Our method uses only two frames as input to the MotionNet to predict the motion between two frames, and then concatenates 4 consecutive predictions to get the result for the 5-frame snippet. In contrast to previous methods, our method also uses stereo information. We can see that the rigid alignment module significantly improves the pose estimation (Ego+align vs. Ego-motion). After training with the flow consistency and rigid alignment modules, ours (full) model further improves and reaches the performance of state-of-the-art methods. 

The other metric is proposed in \cite{zhan2018unsupervised} which measures the average translational and rotational errors for all sub-sequences of length (100, 200, ..., 800). We concatenate all of two frames estimations together for the entire sequence without any post-processing. The results are shown in Table. \ref{tab:odo}.  Our full method outperforms \cite{zhan2018unsupervised} by a large margin, but still falls behind ORB-SLAM (Full) which does bundle adjustments. 

\begin{table}
\centering
\fontsize{7.5}{10}\selectfont
\begin{tabular}{ c | c c  c  c }
   \hlineB{3}
    Method                 & Pixel Acc. & Mean Acc. & Mean IoU & f.w. IoU  \\
    \hline
    \cite{yang2018every} & 0.89  & 0.75 & 0.52 & 0.87  \\
    Ours (Full)            & \bf{0.90} & \bf{0.82} & \bf{0.56} & \bf{0.88}\\ 
  \hlineB{3}  
\end{tabular}
\caption{Motion segmentation evaluation. The metrics are pixel accuracy, mean pixel accuracy, mean IoU,  and frequency weighted IoU. }
\label{tab:mask}
\end{table}

\textbf{Motion Segmentation Evaluation} 
The motion segmentation task is evaluated using the object map provided by the KITTI 2015 dataset. The objects labeled in the object map are treated as moving foregrounds, while all of the remaining pixels are treated as static backgrounds.The metrics follow the ones used in \cite{yang2018every}. We achieve modest improvements in the mean pixel accuracy and mean IoU metrics as shown in Table. \ref{tab:mask}. The qualitative results of our estimated motion segmentation can be seen in Fig. \ref{fig:vis}.

%

\textbf{Ablation Study} The ablation study has already been presented and discussed in previous sections. The comparisons between Ours (Ego-motion), Ours (Ego+align) and Ours (Full) demonstrate the effectiveness of our proposed rigid alignment module and flow consistency module across various tasks of depth, optical flow and odometry estimations. 

\section{Discussions}
In summary, by mutually leveraging stereo and temporal information, and treating the learning of depth and optical flow as a whole, our proposed method shows substantial improvements on unsupervised learning of depth, optical flow, and motion segmentation on the KITTI dataset. However, there is still a need of future works to continue the improvement.

First, the motion segmentation results are still not good enough for practical usage (see supplementary for more examples). More cues for motion segmentation like the ones used in \cite{taniai2017fast} might be useful.

Second, due to the limitations of our motion segmentation, we actually did not fully utilize the accurate rigid flow in our optical flow task (see the performance gap between Ours (Ego+align) and Ours (Full) in the rigid regions). If we were able to get a better motion segmentation mask, the KITTI 2015 optical flow task could be further improved. 

Finally, although our method handles the moving object explicitly, it still assumes that the majority of the scene is static. More work needs to be done to make it suitable for highly dynamic scenes.

{\small
\bibliographystyle{aaai}
\bibliography{egbib}

\begin{thebibliography}{}

\bibitem[\protect\citeauthoryear{Behl \bgroup et al\mbox.\egroup
  }{2017}]{behl2017bounding}
Behl, A.; Jafari, O.~H.; Mustikovela, S.~K.; Alhaija, H.~A.; Rother, C.; and
  Geiger, A.
\newblock 2017.
\newblock Bounding boxes, segmentations and object coordinates: How important
  is recognition for 3d scene flow estimation in autonomous driving scenarios?
\newblock In {\em International Conference on Computer Vision}.

\bibitem[\protect\citeauthoryear{Besl and McKay}{1992}]{besl1992method}
Besl, P.~J., and McKay, N.~D.
\newblock 1992.
\newblock Method for registration of 3-d shapes.
\newblock In {\em Sensor Fusion IV: Control Paradigms and Data Structures},
  volume 1611,  586--607.
\newblock International Society for Optics and Photonics.

\bibitem[\protect\citeauthoryear{Chang and Chen}{2018}]{chang2018pyramid}
Chang, J.-R., and Chen, Y.-S.
\newblock 2018.
\newblock Pyramid stereo matching network.
\newblock In {\em Proceedings of the IEEE Conference on Computer Vision and
  Pattern Recognition},  5410--5418.

\bibitem[\protect\citeauthoryear{DeSouza and Kak}{2002}]{desouza2002vision}
DeSouza, G.~N., and Kak, A.~C.
\newblock 2002.
\newblock Vision for mobile robot navigation: A survey.
\newblock {\em IEEE transactions on pattern analysis and machine intelligence}
  24(2):237--267.

\bibitem[\protect\citeauthoryear{Garg \bgroup et al\mbox.\egroup
  }{2016}]{garg2016unsupervised}
Garg, R.; BG, V.~K.; Carneiro, G.; and Reid, I.
\newblock 2016.
\newblock Unsupervised cnn for single view depth estimation: Geometry to the
  rescue.
\newblock In {\em European Conference on Computer Vision},  740--756.
\newblock Springer.

\bibitem[\protect\citeauthoryear{Geiger, Lenz, and
  Urtasun}{2012}]{geiger2012we}
Geiger, A.; Lenz, P.; and Urtasun, R.
\newblock 2012.
\newblock Are we ready for autonomous driving? the kitti vision benchmark
  suite.
\newblock In {\em Computer Vision and Pattern Recognition (CVPR), 2012 IEEE
  Conference on},  3354--3361.
\newblock IEEE.

\bibitem[\protect\citeauthoryear{Godard, Mac~Aodha, and
  Brostow}{2017}]{godard2016unsupervised}
Godard, C.; Mac~Aodha, O.; and Brostow, G.~J.
\newblock 2017.
\newblock Unsupervised monocular depth estimation with left-right consistency.
\newblock In {\em CVPR}, volume~2, ~7.

\bibitem[\protect\citeauthoryear{Ilg \bgroup et al\mbox.\egroup
  }{2017}]{ilg2016flownet}
Ilg, E.; Mayer, N.; Saikia, T.; Keuper, M.; Dosovitskiy, A.; and Brox, T.
\newblock 2017.
\newblock Flownet 2.0: Evolution of optical flow estimation with deep networks.
\newblock In {\em IEEE Conference on Computer Vision and Pattern Recognition
  (CVPR)}, volume~2.

\bibitem[\protect\citeauthoryear{Jason, Harley, and
  Derpanis}{2016}]{jason2016back}
Jason, J.~Y.; Harley, A.~W.; and Derpanis, K.~G.
\newblock 2016.
\newblock Back to basics: Unsupervised learning of optical flow via brightness
  constancy and motion smoothness.
\newblock In {\em Computer Vision--ECCV 2016 Workshops},  3--10.
\newblock Springer.

\bibitem[\protect\citeauthoryear{Kingma and Ba}{2014}]{kingma2014adam}
Kingma, D., and Ba, J.
\newblock 2014.
\newblock Adam: A method for stochastic optimization.
\newblock {\em arXiv preprint arXiv:1412.6980}.

\bibitem[\protect\citeauthoryear{Li \bgroup et al\mbox.\egroup
  }{2017}]{li2017undeepvo}
Li, R.; Wang, S.; Long, Z.; and Gu, D.
\newblock 2017.
\newblock Undeepvo: Monocular visual odometry through unsupervised deep
  learning.
\newblock {\em arXiv preprint arXiv:1709.06841}.

\bibitem[\protect\citeauthoryear{Mahjourian, Wicke, and
  Angelova}{2018}]{mahjourian2018unsupervised}
Mahjourian, R.; Wicke, M.; and Angelova, A.
\newblock 2018.
\newblock Unsupervised learning of depth and ego-motion from monocular video
  using 3d geometric constraints.
\newblock In {\em Proceedings of the IEEE Conference on Computer Vision and
  Pattern Recognition}.

\bibitem[\protect\citeauthoryear{Meister, Hur, and Roth}{2018}]{Meister2018UUL}
Meister, S.; Hur, J.; and Roth, S.
\newblock 2018.
\newblock {UnFlow}: Unsupervised learning of optical flow with a bidirectional
  census loss.
\newblock In {\em AAAI}.

\bibitem[\protect\citeauthoryear{Menze and Geiger}{2015}]{menze2015object}
Menze, M., and Geiger, A.
\newblock 2015.
\newblock Object scene flow for autonomous vehicles.
\newblock In {\em Proceedings of the IEEE Conference on Computer Vision and
  Pattern Recognition},  3061--3070.

\bibitem[\protect\citeauthoryear{Ranjan \bgroup et al\mbox.\egroup
  }{2018}]{ranjan2018adversarial}
Ranjan, A.; Jampani, V.; Kim, K.; Sun, D.; Wulff, J.; and Black, M.~J.
\newblock 2018.
\newblock Adversarial collaboration: Joint unsupervised learning of depth,
  camera motion, optical flow and motion segmentation.
\newblock {\em arXiv preprint arXiv:1805.09806}.

\bibitem[\protect\citeauthoryear{Ren \bgroup et al\mbox.\egroup
  }{2017}]{ren2017unsupervised}
Ren, Z.; Yan, J.; Ni, B.; Liu, B.; Yang, X.; and Zha, H.
\newblock 2017.
\newblock Unsupervised deep learning for optical flow estimation.
\newblock In {\em AAAI},  1495--1501.

\bibitem[\protect\citeauthoryear{Sun \bgroup et al\mbox.\egroup
  }{2017}]{sun2017pwc}
Sun, D.; Yang, X.; Liu, M.-Y.; and Kautz, J.
\newblock 2017.
\newblock Pwc-net: Cnns for optical flow using pyramid, warping, and cost
  volume.
\newblock {\em arXiv preprint arXiv:1709.02371}.

\bibitem[\protect\citeauthoryear{Taniai, Sinha, and
  Sato}{2017}]{taniai2017fast}
Taniai, T.; Sinha, S.~N.; and Sato, Y.
\newblock 2017.
\newblock Fast multi-frame stereo scene flow with motion segmentation.
\newblock In {\em Computer Vision and Pattern Recognition (CVPR), 2017 IEEE
  Conference on},  6891--6900.
\newblock IEEE.

\bibitem[\protect\citeauthoryear{Tsai, Yang, and Black}{2016}]{tsai2016video}
Tsai, Y.-H.; Yang, M.-H.; and Black, M.~J.
\newblock 2016.
\newblock Video segmentation via object flow.
\newblock In {\em Proceedings of the IEEE Conference on Computer Vision and
  Pattern Recognition},  3899--3908.

\bibitem[\protect\citeauthoryear{Vedula \bgroup et al\mbox.\egroup
  }{1999}]{vedula1999three}
Vedula, S.; Baker, S.; Rander, P.; Collins, R.; and Kanade, T.
\newblock 1999.
\newblock Three-dimensional scene flow.
\newblock In {\em Computer Vision, 1999. The Proceedings of the Seventh IEEE
  International Conference on}, volume~2,  722--729.
\newblock IEEE.

\bibitem[\protect\citeauthoryear{Wang and Buenaposada}{2018}]{wang2018learning}
Wang, C., and Buenaposada, J.~M.
\newblock 2018.
\newblock Learning depth from monocular videos using direct methods.
\newblock In {\em Proceedings of the IEEE Conference on Computer Vision and
  Pattern Recognition}.

\bibitem[\protect\citeauthoryear{Wang \bgroup et al\mbox.\egroup
  }{2018}]{wang2018occlusion}
Wang, Y.; Yang, Y.; Yang, Z.; Zhao, L.; and Xu, W.
\newblock 2018.
\newblock Occlusion aware unsupervised learning of optical flow.
\newblock In {\em Proceedings of the IEEE Conference on Computer Vision and
  Pattern Recognition},  4884--4893.

\bibitem[\protect\citeauthoryear{Wulff, Sevilla-Lara, and
  Black}{2017}]{wulff2017optical}
Wulff, J.; Sevilla-Lara, L.; and Black, M.~J.
\newblock 2017.
\newblock Optical flow in mostly rigid scenes.
\newblock In {\em The IEEE Conference on Computer Vision and Pattern
  Recognition (CVPR)}.

\bibitem[\protect\citeauthoryear{Yang \bgroup et al\mbox.\egroup
  }{2017}]{yang2017unsupervised}
Yang, Z.; Wang, P.; Xu, W.; Zhao, L.; and Nevatia, R.
\newblock 2017.
\newblock Unsupervised learning of geometry with edge-aware depth-normal
  consistency.
\newblock {\em arXiv preprint arXiv:1711.03665}.

\bibitem[\protect\citeauthoryear{Yang \bgroup et al\mbox.\egroup
  }{2018}]{yang2018every}
Yang, Z.; Wang, P.; Wang, Y.; Xu, W.; and Nevatia, R.
\newblock 2018.
\newblock Every pixel counts: Unsupervised geometry learning with holistic 3d
  motion understanding.
\newblock {\em arXiv preprint arXiv:1806.10556}.

\bibitem[\protect\citeauthoryear{Yin and Shi}{2018}]{yin2018geonet}
Yin, Z., and Shi, J.
\newblock 2018.
\newblock Geonet: Unsupervised learning of dense depth, optical flow and camera
  pose.
\newblock In {\em Proceedings of the IEEE Conference on Computer Vision and
  Pattern Recognition (CVPR)}, volume~2.

\bibitem[\protect\citeauthoryear{Zhan \bgroup et al\mbox.\egroup
  }{2018}]{zhan2018unsupervised}
Zhan, H.; Garg, R.; Weerasekera, C.~S.; Li, K.; Agarwal, H.; and Reid, I.
\newblock 2018.
\newblock Unsupervised learning of monocular depth estimation and visual
  odometry with deep feature reconstruction.
\newblock In {\em Proceedings of the IEEE Conference on Computer Vision and
  Pattern Recognition},  340--349.

\bibitem[\protect\citeauthoryear{Zhou \bgroup et al\mbox.\egroup
  }{2017a}]{zhou2017unsupervisedstereo}
Zhou, C.; Zhang, H.; Shen, X.; and Jia, J.
\newblock 2017a.
\newblock Unsupervised learning of stereo matching.
\newblock In {\em International Conference on Computer Vision}.

\bibitem[\protect\citeauthoryear{Zhou \bgroup et al\mbox.\egroup
  }{2017b}]{zhou2017unsupervised}
Zhou, T.; Brown, M.; Snavely, N.; and Lowe, D.~G.
\newblock 2017b.
\newblock Unsupervised learning of depth and ego-motion from video.
\newblock In {\em CVPR}, volume~2, ~7.

\end{thebibliography}
}

\end{document}